\documentclass[letterpaper, 10 pt, conference]{ieeeconf}  
\let\labelindent\relax                                                        

\IEEEoverridecommandlockouts                              
\usepackage{etoolbox}
\patchcmd{\keywords}{Index Terms}{Keywords}{}{}
\usepackage{amsmath}
\usepackage{float}
\usepackage{enumitem}
\usepackage{tabularx}
\usepackage{stmaryrd}
\usepackage{makecell}
\usepackage{graphicx} 

\title{\LARGE \bf
Effectiveness of Hierarchical Softmax in Large Scale Classification Tasks
}

\author{Abdul Arfat Mohammed$^{1}$ and Venkatesh Umaashankar$^{2}$
\thanks{$^{1}$Abdul Arfat Mohammed - Intern at Ericsson R\&D, Chennai.\newline
        {\tt\small  E-mail : arfat999@gmail.com}}%
\thanks{$^{2}$Venkatesh Umaashankar - Researcher at Ericsson R\&D, Chennai.\newline
        {\tt\small  E-mail : venkatesh.u@ericsson.com}}%
        }

\begin{document}
\maketitle
\thispagestyle{empty}
\pagestyle{empty}

\begin{abstract}
Typically, $Softmax$ is used in the final layer of a neural network to get a probability distribution for output classes. But the main problem with $Softmax$ is that it is computationally expensive for large scale data sets with large number of possible outputs. To approximate class probability efficiently on such large scale data sets we can use $Hierarchical\ Softmax$. LSHTC datasets were used to study the performance of the $Hierarchical\ Softmax$. LSHTC datasets have large number of categories. In this paper we evaluate and report the performance of normal $Softmax$ Vs $Hierarchical\ Softmax$ on LSHTC datasets. This evaluation used macro f1 score as a performance measure. The observation was that the performance of $Hierarchical\ Softmax$ degrades as the number of classes increase.
\end{abstract}

\section{INTRODUCTION}\label{intro}

\begin{figure}[H]
  \centering
  \includegraphics[scale=0.45]{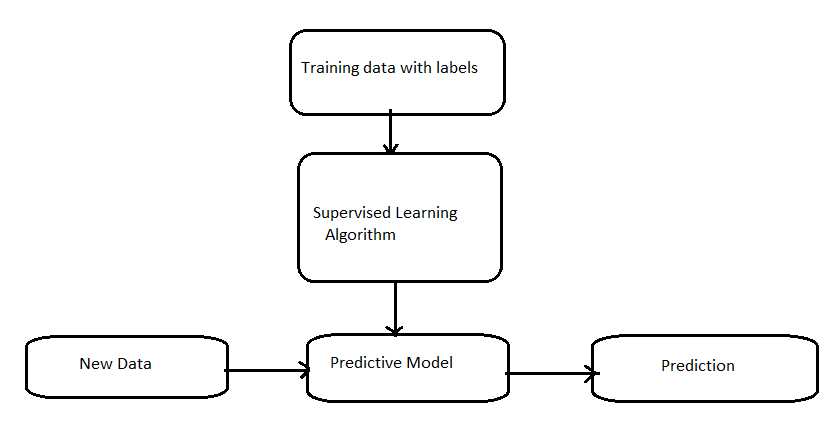} 
  \caption{\textmd{Classification Block Diagram}}
  \label{cly}
\end{figure}

\textbf{Classification} is the problem of finding to which of a set of labels a new 
instance belongs, on the basis of a training data set containing instances 
whose label membership is already known. Supervised in the Figure \ref{cly} refers to the fact that the instances of training set have their label already known.
A simple example would be like classifying fruits to a type based on past fruits dimension data. Suppose, say we want to find if a fruit is apple (or) banana based on the height of the fruit. And assume our previous data suggests that fruits that were apples had their height in range of $9-11$ cm and that of banana was in range $13-20$ cm. Learning on above data is supervised because fruits that had height in range of $9-11$ cm were apples and between $13-20$ cm were bananas. Hence, our data is labelled. Our learning algorithm could be as simple as to see the new fruit’s height and then check under which fruit’s height range it falls. So, using the above learning algorithm on our training data  (with labels) we make a predictive model which can be used to classify new instances. Suppose, say our new fruit (new data) which is to be classified has a height of $14$ cm, we simply classify that as a banana (prediction). 

\textbf{Loss Function}, also known as objective function is a evaluation measure of the model, typically lower the loss value better the model is . There are variety of loss functions such as cross entropy, hinge, mean squared error etc. Each loss function has its own pros and cons. The chosen loss function not only determines the model performance but also the run time and complexity.\textbf{ Activation functions }typically take some set of inputs and map them into a non-linear space. Activation functions can also give us a probability distribution (i.e., the sum of probabilities of all nodes in the final layer is 1) for the final layer in a neural network. Some examples for activation functions are $Softmax, Sigmoid, Tanh$ etc. Generally, In deep learning $Softmax$ is used as an activation function on the final layer to return class probabilities for prediction purposes.

\textbf{Text Classification} is the task of assigning predefined categories to free-text documents. An example would be like classifying whether an email is spam or not. Similar to the above fruit classification we can train a supervised learning algorithm on a corpus of labelled emails (emails which are already categorized as Spam or not Spam) to make a prediction model which then can be used to predict new emails as they arrive either as Spam or not Spam. In Multi-class classification there are more than two categories available and the new instance can belong to only one of those categories. An example would be like classifying whether a fruit is an apple (or)  a mango (or) a banana. In Multi-label classification an instance can belong to more than one category at the same time. An example would be like a document may be about politics, sports and religion at the same time.

\textbf{In Large Scale Classification Tasks,} we try to classify million (or) more instances into thousands (or) more categories which may be both multi-class and multi-label in nature. Example: Classifying the category of a Wikipedia document could be both multi-class and multi-label task.

In this paper, our main \textbf{contributions} are, (1) We prepare custom datasets with various distinct category sizes \textit{n(=10,100,1000,10000)} from LSHTC data sets\cite{lshtc} as explained in the Section \ref{sec:experiment_setup}. (2)  We trained Fasttext models to compare the performance of $Hierarchical\ Softmax$ Vs $Softmax$ in on these data sets using macro f1-score as an evaluation metric. (3) We report our observations and findings in the Section. \ref{sec:results}.

\section{Large Scale Classification}
Large Scale Classification, typically involves dealing with millions of documents and thousands of categories. One such task is Large Scale Hierarchical Text Classification \textbf{(LSHTC)}. In this paper, we used the data set corresponding to 4th edition of the Large Scale Hierarchical Text Classification (LSHTC) Challenge \cite{PartalasKBAPGAA15} for evaluating the performance of using $Hierarchical\ Softmax$ as activation function instead of plain $Softmax$. The LSHTC Challenge is a hierarchical text classification competition, using very large datasets. The challenge is based on a large dataset created from Wikipedia. The dataset is multi-class, multi-label and hierarchical. 

\textbf{FastText} \cite{fasttext} is an open-source, free, lightweight library that allows users to learn text representations and text classifiers. Also, FastText provides an implementation of $Hierarchical\ Softmax$ based activation function for text classification purposes \cite{joulin2016bag}. We used implementation of FastText for conducting the experiments in this paper.

\subsection{Softmax}
$Softmax$ \cite{bendersky_2016} function takes an \textit{N}-dimensional vector of arbitrary real values and produces another \textit{N}-dimensional vector with real values in the range $(0, 1)$ that add up to $1$, i.e., $Softmax$ turns the \textit{N}-dimensional vector into a probability distribution which can be used for prediction purposes.

Mathematical formula:
\begin{equation}
S_i=\frac{e^{a_i}}{\sum_{j=1}^{N} e^{a_j}}
\label{eqq}
\end{equation}

here, $S_i$ is the $Softmax$ output for $i^{th}$ value in our input vector of size \textit{N}. Our Input vector is $[a_1,a_2,......a_N]$.

$S_i$ is always positive i.e., $S_i>0$ because of exponents. As the numerator appears in the denominator summed up with some other positive numbers, $S_i$  $<$ 1. Hence, this property enables us to derive a probability distribution for the classes in a classification. Typically, the $Softmax$ is used along with the cross-entropy loss function in a neural network based classifier. 

\begin{equation}
\label{eq:softmax}
Softmax([1,2,3,4]) = [0.03,0.08,0.24,0.64]
\end{equation}

Equation \ref{eq:softmax} shows an example of 
applying $Softmax$ function on a four element input vector. The order of elements by relative size is preserved, and they add up to $1$. Intuitively, the $Softmax$ activation function is a soft version of the maximum function.

For large numbers (positive or negative), computing $Softmax$ may cause numeric instability due to the exponentiation. To handle this we can normalize the input vector. For detailed description on how $Softmax$ is used in learning vector representation for words, refer to continuous bag-of-word model (CBOW) introduced in \cite{mikolov2013efficient} and skip-gram model  introduced in \cite{mikolov2013distributed}. For further details on how the parameter learning takes place in word2vec, refer \cite{rong2014word2vec}. 

A key problem in using $Softmax$ for Large Scale Classification is that computation becomes expensive because of the normalizing sum in the denominator term of the Equation \ref{eqq}. This not only affects the forward pass, but also slows down the back propagation.  To solve this problem, an intuition is to limit the number of output vectors that must be updated per training instance. One way to avoid this problem is to use $Hierarchical\ Softmax$. 
\subsection{Hierarchical Softmax}
$Hierarchical\ Softmax$ is an efficient way of computing $Softmax$ \cite{morin2005hierarchical}. Binary trees are used to represent all categories in the
output dictionary $(V)$ in this model. The leaf units of this binary tree represent the categories in the output dictionary $(V)$. For each leaf unit we have a unique path from the root and the probability of a category is estimated using this path. As it is a binary tree it is obvious that there are $(V-1)$ intermediate nodes. Here, the inner nodes represent internal parameters $(probability$ $mass)$. Each intermediate node explicitly represents the relative probabilities of its child nodes. The idea behind decomposing the output layer to a binary tree was to reduce the complexity to obtain probability distribution from $O(V)$ to $O(\log (V)).$ 
Below Figure \ref{btree} is an example of Hierarchical binary tree (figure is from \cite{rong2014word2vec}):
\begin{figure}[H]
\centering
\includegraphics[scale=0.5]{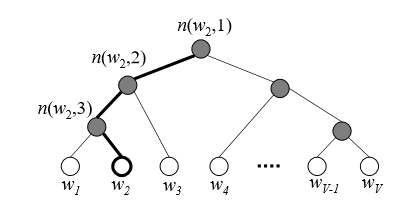}
\caption{\textmd{An example binary tree for the $Hierarchical\ Softmax$ model}} 
\label{btree}
\end{figure}
The leaf nodes represent the categories and the inner nodes represent probability mass. The highlighted nodes and edges make a path from root to the above example leaf node $w_2$. Here, length of the path $L(w_2)$= 4.
$n(w,j)$ means the $j^{th}$ node on the path from root to a leaf node $w$.
In $Hierarchical\ Softmax$ model, each of the $(V-1)$ intermediate node has an output vector $v'_{n(w, j)}$ instead of output vector representation for words. And the probability of a category $w$ being the output class will be as follows:
\begin{equation}
\small p(w=w_O) = \small\prod_{j=1}^{L(w)-1}  \sigma\left(\llbracket n(w,j+1) = ch(n(w,j))\rrbracket \cdot v'^T_{n(w,j)}h \right)
\end{equation}
here, $w_O$ is the actual output category. $\sigma$ is the sigmoid function.
$\llbracket x\rrbracket$: is a special function defined as:
\begin{equation}
\llbracket x\rrbracket =
\begin{cases} 
      1 & $if $x $ is true$; \\
      -1 & otherwise. 
   \end{cases}
\end{equation}
$h$ is the output of the hidden layer.
$v'_{n(w, j)}$ is the vector representation of the intermediate node $n(w, j)$ and 
$ch(n(w,j))$ is the left child of $n(w,j)$.

Probabilities at each intermediate node in the path from root to the output category are required to compute the probability of the output category. To achieve this, at each intermediate node we must assign the probabilities for going right and going left.

We define the probability of going left at an intermediate node $n$ as follows:
\begin{equation}
p(n,left) = \sigma\left(v'^T_n \cdot h\right)
\end{equation}
And the probability of going right at the same node $n$ will obviously be:
\begin{equation}
p(n,right) = 1 - \sigma\left(v'^T_n \cdot h\right) = \sigma\left(-v'^T_n \cdot h\right)
\end{equation}

We can calculate the probability for the category $w_2$ in the Figure \ref{btree} as follows:
\begin{align*}
p(w_2 = w_O) &= p(n(w_2,1),left)\cdot p(n(w_2,2),left)\\
&\quad\cdot p(n(w_2,3),right)\\
&= \sigma\left(v'^T_{n({w_2},1)} h\right) \cdot \sigma\left(v'^T_{n({w_2},2)} h\right)\\
&\quad \cdot\sigma\left(-v'^T_{n({w_2},3)} h\right)
\end{align*}

For the detailed description on how the vector representations for the inner nodes are learned, refer \cite{rong2014word2vec}.

We can also check that sum of calculated probabilities for all the words in the vocabulary add up to $1$, which is the intuition for using $Softmax$ in the first place, but $Hierarchical\ Softmax$ does this in a faster manner.
\begin{equation}
\label{eq1}
\sum_{i=1}^V p(w_i = w_O) = 1
\end{equation}
$Hierarchical\ Softmax$ is a well-defined multinomial distribution among all output categories. This implies that the cost for computing the loss function and its gradient will be proportional to the number of nodes $(V)$ in the intermediate path between root node and the output node, which on average is no greater than  $\log (V)$.
The performance also depends on the Hierarchical tree structure used, having said that binary Huffman tree \cite{mikolov2013efficient} \cite{joulin2016bag} is expected to optimize tree for faster training.

\subsection{Evaluation}
LSHTC used macro f1 score as the performance metric for the evaluation criterion. To make our study comparable to their evaluation, we used the same macro f1 score as our performance metric. The description of the macro f1 score is detailed in the method subsection \ref{meth}.

\section{Experimental Setup} \label{sec:experiment_setup}
As mentioned in previous sections, we used the labeled data available in LSHTC for conducting the experiments in this paper. The entire process of creating custom data sets for top \textit{n} labels where, \textit{n(=10,100,1000,10000)} is described in method subsection \ref{meth} below. So, for four different values of \textit{n}, four separate data sets were created from the LSHTC labeled data set. The specifications of the machine used for this experiment are as follows:
\begin{itemize}
\item \textbf{CPU :} Intel® Xeon® Processor E5-2650 v4 30M Cache, 2.20 GHz, 12 Cores, 24 Threads
\item \textbf{RAM :} 250 GB
\item \textbf{OS  :} CentOS 7
\end{itemize}
\subsection{Dataset}\label{ds}
The LSHTC \cite{lshtc} labeled data was in preprocessed form.
This LSHTC is a multi-class and multi-label problem. This data set has  $3,25,056$ categories to be exact and $23,65,437$ instances of labeled data is available. The format of each data file follows the libSVM format. Each line corresponds to a sparse document vector and has the following format:
\[label,\enspace label,\enspace label ... feat:value ... feat:value\]
\textbf{ label} is an integer and corresponds to the category to which the document vector belongs. Each document vector may belong to more than one category. The pair $feat:value$ corresponds to a non-zero feature with index $feat$ and value $value$.\textbf{ feat }is an integer representing a term and \textbf{value} is a double that corresponds to the weight \textit{(tf)} of the term in the document.
For example:
\[545,\enspace 32\enspace8:1\enspace18:2\]
corresponds to a document vector whose features are all zeros except feature number $8$ (with value $1$) and feature number $18$ (with value $2$). This document vector belongs to categories $545$ and $32$. Each feature number is associated to a stemmed word.

\subsection{Method}\label{meth}
We used Fasttext for the text classification process with the following hyper parameters:
\begin{table}[H]
\centering
\begin{tabular}{||c c||}
 \hline
 Hyper Parameter Name & Hyper Parameter Value \\ [0.5ex] 
 \hline\hline
 dim (dimension for word)& 200 \\ 
 epoch & 100\\
 lr (learning rate)& 0.25 \\
 loss (loss function)& hs\\ [1ex] 
 \hline
\end{tabular}
\caption{\textmd{Hyper Parameter Values used for Fasttext }}
\label{table:5}
\end{table}here, hs is $Hierarchical\ Softmax$.

The hierarchy file provided in LSHTC was not used. The labeled data available in LSHTC data set was split into $70\%$ for training and $30\%$ for testing (data was shuffled before splitting). The macro f1 score was used as performance measure. The evaluation was done for top \textit{n} labels each time \textit{(n=10,100,1000,10000)}. The top \textit{n} signifies the top \textit{n} labels occurring in the LSHTC training data. 

By default FastText predicts one label per sample. But, we can predict more labels per sample by giving a number as an attribute to FastText's predict function.

As LSHTC is a multi class and multi label problem, we had to decide how many labels to predict for each sample. For this we calculated the average number of labels per doc from the LSHTC training file for each \textit{n}. So, the evaluation is done for \textit{n} values of \textit{10,100,1000,10000} by following the below steps:

1. Find the top \textit{n} labels occurring in the LSHTC labeled data.

2. Based on the above top \textit{n} labels, create a new file which has samples for only those classes from the LSHTC labeled data. If a doc has a label from top \textit{n} labels and other labels not from top \textit{n} labels then, instead of discarding the doc, only the labels which are not in top \textit{n} are discarded and we keep the doc to get as many number of samples for the training process. Here, we strip the term frequencies from the instances and prepend $feat$ with \textit{'w'} to treat it as a word (as discussed in data set sub section \ref{ds}) to prepare the input data in FastText format. For example, consider \[545,\enspace 32\enspace8:1\enspace18:2\] we will format this instance as follows:
\[\_\_label\_\_545\enspace\_\_label\_\_32\enspace w8\enspace w18\]

3. After creating the new data, we find the number of average labels available for each doc and use the rounded number of that average as an argument for predict method of FastText, say it is \textit{x}.

4. The new data is then shuffled and split into $70\%$ train set and $30\%$ test set.

5. Now we use FastText to train on $70\%$ of the data using the hyper parameters shown in Table \ref{table:5}.

6. We then predict \textit{x} labels per doc (\textit{x} is from step 3) using fasttext.

7. For evaluation we use macro f1 score ($MaF$) as follows:
	\begin{equation}
	    MaF=\frac{2*MaP*MaR}{MaP+MaR}
	\end{equation}
   Here, $MaP$ is macro precision and $MaR$ is macro recall.
   \begin{equation}
   MaP=\frac{\sum_{i=1}^{|C|}\frac{tp_{c_i}}{tp_{c_i}+fp_{c_i}}}{|C|}
   \end{equation}
   \begin{equation}
   MaR=\frac{\sum_{i=1}^{|C|}\frac{tp_{c_i}}{tp_{c_i}+fn_{c_i}}}{|C|}
   \end{equation}
where, \textit{C} is set of classes, $tp_{c_i}, fn_{c_i} $and $fp_{c_i}$ are the true positives, false negatives and false positives respectively for class $c_i$.
The above process is repeated for each value of \textit{n}.
\section{Results}
\label{sec:results}
The following tables show the results obtained for various top \textit{n} labels in LSHTC training data.
All macro precision, macro recall, macro f1 scores in the Tables \ref{table:2} and \ref{table:3} are rounded to two decimal places.
\begin{table}[H]
\centering
\begin{tabular}{||c c c ||}
 \hline
 Top \textit{n} labels & avg labels per doc& labels predicted per doc \\ [0.5ex] 
\hline\hline
 10 & 1.2 & 1 \\ 
 100 & 1.4 & 1  \\
 1000 & 1.9 & 2  \\
 10000 & 2.4 & 2  \\ [1ex] 
\hline
\end{tabular}
\caption{\textmd{Results obtained for various top \textit{n} labels data set} }
\label{table:1}
\end{table}
\begin{table}[H]
\centering
\begin{tabular}{|| c c c c||}
\hline
 Top \textit{n} labels & macro precision & macro recall & macro f1 \\ [0.5ex] 
 \hline\hline
 10 &0.77 & 0.45 & 0.54 \\ 
 100 & 0.51 & 0.30 & 0.35 \\
1000 &0.34 & 0.34 & 0.32 \\
  10000 &0.25 & 0.22 & 0.21 \\
  [1ex] 
\hline
\end{tabular}
\caption{\textmd{Scores obtained for various top \textit{n} labels using $Hierarchical\ Softmax$}}
\label{table:2}
\end{table}
\begin{table}[H]
\centering
\begin{tabular}{|| c c c c||}
\hline
  Top \textit{n} labels & macro precision & macro recall & macro f1 \\ [0.5ex] 
 \hline\hline
 10 &0.82 & 0.48 & 0.58 \\ 
 100 & 0.69 & 0.40 & 0.47 \\
1000 &0.55 & 0.46 & 0.47 \\
  10000 &0.49 & 0.37 & 0.38 \\
  [1ex] 
\hline
\end{tabular}
\caption{\textmd{Scores obtained for various top \textit{n} labels using $Softmax$}}
\label{table:3}
\end{table}
\begin{table}[H]
\centering
\begin{tabular}{||c c c ||}
 \hline
 Top \textit{n} labels & $Hierarchical\ Softmax$ & $Softmax$ \\ [0.5ex] 
 \hline\hline
 10 & 1 min & 1 min\\ 
 100 & 1 min & 5 min\\
 1000 & 2 min & 51 min\\
 10000 & 5 min& 15 Hr 11 min\\ [1ex] 
 \hline
\end{tabular}
\caption{\textmd{Training times for models when $Hierarchical\ Softmax$ and $Softmax$ are used }}
\label{table:4}
\end{table}
Based on top \textit{n(=10,100,1000,10000)} labels, data sets for each \textit{n} are created separately as mentioned in the method sub section \ref{meth}. Average labels per doc are calculated from created data set (different for each of top \textit{n} labels). Labels predicted per doc is the number of labels we are predicting (using fasttext predict option) for each doc in test set of our created data set (different for each of top \textit{n} labels). The macro precision, macro recall, macro f1 scores are calculated as described in the method sub section \ref{meth}. 
The Figure \ref{g1} below, shows how the macro f1 score (for top \textit{n} labels) varies when $Hierarchical\ Softmax$ and $Softmax$ are used on the final layer of a neural network. We can also see that the macro f1 score decreases as the number of labels increase.
\begin{figure}[H]
\centering
\includegraphics[scale=0.5]{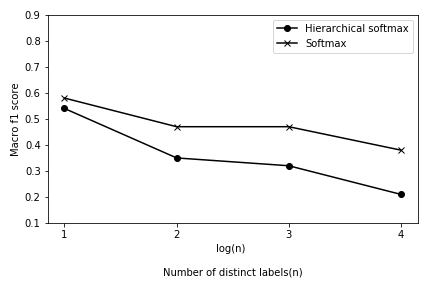}
\caption{\textmd{Comparison of macro f1 scores when $Hierarchical\ Softmax$ and $Softmax$ are used}}
\label{g1}
\end{figure}
In the Figure \ref{g1} above, we used the $\log_{10}$ scale for the top \textit{n} classes because as the size of the labels grows the fall in the macro f1 score was increasing and it was not intuitive enough to view on scale where \textit{n} ranges from $10$ to $10000$, hence the graph was plotted with $\log_{10}(n)$ on the $x-axis$ and with the macro f1 score on the $y-axis$.

From Tables \ref{table:2}, \ref{table:3} and \ref{table:4}, It is evident that $Hierarchical\ Softmax$ indeed improves training speed when compared to $Softmax$, but at the cost of being less accurate.
\section{Conclusion}
In this paper, we compared the performance of $Hierarchical\ Softmax$ and $Softmax$ in large scale classification tasks on LSHTC data. Our conclusions are as follows:
(1) $Softmax$ performs better when compared to $Hierarchical\ Softmax$ in large scale classification tasks. (2) $Hierarchical\ Softmax$ is indeed faster when compared to $Softmax$ in training a model.

In our future work, we will expand our study in the following areas: (1) Evaluate using a data set that is in raw text form and not in pre-processed form. (2) Use better model to find the optimal number of labels to predict per doc instead of simple averaging based approach. (3) Combine FastText embeddings with a Recurrent Neural Network (RNN) to treat the hierarchical classification as a sequence prediction problem.

\bibliographystyle{IEEEtran}
\bibliography{IEEEabrv,latestpaper} 
	
\end{document}